\documentclass{article}

\usepackage{spconf,graphicx}
\graphicspath{ {./images/} }

\title{Towards Precision Characterization of Communication Disorders \\ 
using Models of Perceived Pragmatic Similarity}
\name{\small{Nigel G. Ward\(^1\), Andres Segura\(^1\), Georgina Bugarini\(^1\),  Heike Lehnert-LeHouillier\(^2\), Dancheng Liu\(^3\), Jinjun Xiong\(^3\), Olac Fuentes\(^1\)}}
\address{
  \(^1\)University of Texas at El Paso, \(^2\) New Mexico State University, \(^3\) University at Buffalo \\United States}

\begin{document}
\ninept

\maketitle

\begin{abstract}
The diagnosis and treatment of individuals with communication disorders
offers many opportunities for the application of speech technology,
but research so far has not adequately considered: the diversity of conditions,
the role of pragmatic deficits, and the challenges of limited data.
This paper explores how a general-purpose model of perceived
pragmatic similarity may overcome these limitations.  
It explains 
how it might support several use cases for clinicians and clients,
and presents evidence that a simple model can provide value,
and in particular can capture utterance aspects
that are relevant to diagnoses of autism and specific language impairment.
\end{abstract}

\begin{keywords}
precision medicine, dialog behaviors, pretraining, low-resource tasks, autism, atypical behavior
\end{keywords}


\section{Motivations}    \label{sec1}

Communication disorders affect many people's lives, and the potential
for speech technology to help has inspired a much research.
Most successes in this area, however, follow a pattern
\cite{milling2022}, in being limited to categorical diagnoses, limited
to cases where ample and convenient labeled data is available
\cite{linzen-2020-accelerate}, and limited in terms of how much the
models contribute to clients' understanding of their conditions,
and limited in alignment with the concept  of ``precision medicine" \cite{johnson2021precision}, and the aim of deeper modeling of  
conditions with complex etiologies (such as psychiatric disorders
\cite{insel2014nimh,passos2022precision,foltz2023reflections}, where 
an important goal is discover latent factors, 
and eventually ties to genetic, neural,
developmental, environmental, or other root causes).
While such research has produced models that perform well on many
data sets, this pattern is not suitable for many needs, 
and translation to clinical practice has been slow  \cite{martin2024voice}.
We see three general needs.

\smallskip\noindent{\bf Need 1: Precise Diagnosis and Categorization} 

\sloppy
Many individuals with communication disorders have non-stereotypical
conjunctions of symptoms, each perhaps present to varying degrees
\cite{martin2024voice}.  For example, someone diagnosed with autistism may
have some characteristics mildly or not at all, and may also have some
elements of anxiety or other conditions.  There are also many people
with subclinical traits, some of whom may still wish to better
understand their condition and learn to communicate better.  Such
people can potentially benefit from information beyond being tagged
with one of a few possible labels.  
While previous work has mostly targeted simplistic diagnosis,
we aim to support more nuanced characterizations.  These could be useful for clinicians and
also for clients seeking to self-monitor, especially for individualized
treatments.
Incidentally, while there are many qualms about automated
rather than human-provided diagnoses, AI involvement in helping
clients better understand their conditions is less likely to be  concerning.

\smallskip\noindent{\bf Need 2: Small-Data Modeling}

Obtaining enough data to build models of communications disorders is
always a challenge, often due to various privacy concerns.  In the
usual research pattern, an assumption of heterogeneity within each
disorder is taken to justify grouping data across many individuals.
This is a convenient way to boost the usable data size, but comes at
the cost of imprecise categorization, as noted above.  Further, many
factors affect language skills, so the reference data should ideally
be limited to speech from people with the same age, gender, demographic,
family environment, dialect, and so on \cite{cohen2016psychometric},
also reducing the amount of relevant data.  Moreover, the increasing
desire to evaluate communicative effectiveness in naturalistic
conditions --- especially in dialog --- rather than via read speech, further
reduces the amount of comparable and good-quality data likely to be
available.  These considerations impel us to find ways to get more out
of limited data.

\smallskip\noindent{\bf Need 3: Consideration of Pragmatic Aspects}

For some communication disorders, especially the neurogenic ones, 
the greatest issue is difficulty in 
correctly conveying or perceiving {\it pragmatic} intents,
and the abilities to use speech functionally and in interaction can be 
the most important aspects for communicative effectiveness and acceptance
by peers \cite{bottema-beutel17}.  
Typical pragmatic functions include, for example, the intent to 
praise, to mark contrast or novelty, to change the topic, to yield the
turn or take it, and so on.  
Phonological, lexical and syntactic deficits are easier to
measure, and have been the focus of most modeling to date, but 
pragmatic behaviors and perceptions also matter,
the latter because, of course, 
social  success depends not only on the intent of the speaker but also on
interlocutors' perceptions.

\section{Modeling Pragmatic Similarity}

This paper explores how  a general-purpose model
of perceived pragmatic similarity might help with these issues. 
In particular, we envisage a core module that takes as input two recorded
utterances and outputs an estimate of their perceived pragmatic
similarity.  While semantic similarity models exist \cite{weiting,besacier2022textless, gehrmann2023repairing,seamless2023}, modeling {\it
  pragmatic} similarity is a new idea \cite{me-interspeech24,heffernan2024}.

This paper also reports preliminary results with a specific model of
pragmatic similarity, a slightly modified version of Segura's  \cite{me-interspeech24}.  
In brief, Segura's model computes, for each utterance, 
the features of a pretrained HuBert model's 24th layer, downselected via a custom feature-selection process to maximize performance.  
The cosine between the feature vectors for the two utterances is then 
taken as the prediction of the similarity.
Compared to a large set of human judgments of
American English utterances \cite{ward-marco},
this model's predictions correlate 0.74 with the judgments, 
which is not far below human inter-annotator agreement. 

\section{Support for New Use Cases}  \label{exemplars-and-outliers}

There are several ways that a pragmatic similarity metric could
serve to address the needs identified above.

\indent {\bf Use Case A: Detecting Atypical Speakers}.  Given a
collection of recordings of people, clinicians might like a tool to
automatically screen for atypical speakers.  

{\bf Use Case B: Finding Similar Speakers}. 
There may be value in identifying a very similar speaker in a dataset.
For example, enabling clients to hear speech from similar others, 
that is, people whose behaviors are
perceived similarly, might help them understand how they themselves
are perceived by other people.  This could help them more vividly
understand their prospects, and also enable them
to match up with appropriate support groups.  
Further, while some clinicians may have enough experience 
to recall how other clients, similar in some way to the current
client, responded to various interventions, novice 
clinicians may lack such experience.  
Thus it could be useful for clinician training or to
provide support for ``looking up'' similar clients via their
recordings in a dataset.

{\bf Use Case C: Finding Typical/Representative Utterances}. While Use
Cases A and B relate to identifying individuals, we can also use a
similarity model to identify utterances of interest.  
In particular, clinicians might appreciate the ability
to quickly find and hear a ``typical'' utterance by the speaker, 
serving as a useful summary of the properties of many
utterances, and thereby enabling faster assessment.

{\bf Use Case D: Finding Comparable Utterances}.  
Clinicians may want to be able to use a specific
utterance from a child of interest as a ``query'' and have the system
retrieve similar utterances from other speakers in the dataset.

{\bf Use Case E:  Identifying Atypical Utterances}.  Clinicians may
also, conversely, be able to benefit from automatic support for
finding {\it atypical} utterances.
In particular, they might want to be able to hear, for a single
speaker, utterances that they have produced that are saliently
non-typical (distant from anything in the reference data), and thus
likely to be possibly perceived negatively by peers.
Further, detection of atypical utterances could be useful for another
reason: as a preprocessing step to classification or to Use Cases A
and B, for the sake of excluding utterances that seem to be outliers,
perhaps due to noise or some unusual transient speaker state, and are
thus likely uninformative.
Alternatively, these atypical utterances might be informative
as children's ``leading-edge" behaviors \cite{rispoli-hadley}.

As a general comment, relevant to Use Cases C, D, and E, 
assessing and diagnosing dialog abilities and behavior patterns
is today labor-intensive.  Speech and language professionals may need to
observe substantial amounts of behavior to make diagnoses, typically
collected in structured ways to support calibrated diagnoses. 
A similarity model may support finding rare but highly informative data.  
For example, it may be that there are ``markers'' 
for breakdowns, and that these markers occur more frequently in more
natural interactions.  Accordingly, we wish to support 
clinicians in finding such especially informative speech segments,
especially for in-the-wild dialog recordings, which may be voluminous.

\section{Preliminary Validation}

Key to all of these use cases is the ability of a model to identify the pragmatically {\it most} similar utterances.
Previous evaluation of Segura's model considered mostly  
overall correlations, but these are heavily swayed
by the basic ability to separate rather similar pairs from completely dissimilar pairs.
We accordingly set out to measure the ability of the model to make precise rankings of similarity among pairs that are all at least fairly similar.   
This ability is necessary for nearest-neighbor models, as illustrated in the next section, 
and for all five of the use cases.
Specifically, we hypothesized that the modified Segura model would be able to discriminate between
the most similar utterances, and those that were not similar or less similar.  
We expected its performance to be far above chance.
To test this ability we designed a task inspired by Use Case D. 

\subsection{Stimuli}

We started with the DRAL corpus, which contains 2893 pragmatically-varied English utterances 
from conversations among 129 college students.  
Ideally, for a direct test of the model, we would have
humans first identify, for each of these utterances, the most similar utterances
in the corpus.  
We would then have been able to directly test the ability of the model, 
given any utterance, to find the most
similar.  However, having subjects listen to thousands of utterance pairs 
would not be realistic. 
Instead, we downsampled, leveraging our model to ensure that very similar utterances were over-represented.

Specifically, we randomly selected 31 reference utterances from the corpus.  
Each of these was then paired with 10 candidates for most-similar, and those 10 pairs were
presented to human subjects for judgments. 
The 10 candidates were chosen using the  Segura model: 
Three were the top-three most similar, according to the model, and, as distractors,
the seven whose similarity levels were at percentiles 99, 97, 95, 90, 80, 60, and 30, according to the model.
In this way we ensured that very similar utterances were over-represented.
All clips were 2-7 seconds long. 
To avoid speaker-identify confounds, each candidate utterance was constrained to be 
by a speaker different from the
speaker of the reference utterances.   

\subsection{Subjects, Instruments, and Procedure}

We recruited nine judges from among the pool of those involved in our recent past studies, 
and through word of mouth, choosing people we knew to be highly sensitive to the nuances of language. 

For each pair, judges rated ``how pragmatically similar are the two clips, 
in terms of the overall feeling, tone, and intent" \cite{ward-marco} on a scale from 1 to 5.
They also recorded their top 3 rankings, which, 
although not really providing additional information, likely encouraged  
extra care with the judgments at the high end of the scale. 

Subjects came in on a Saturday for a 4-hour session.  
Their compensation was 70 dollars plus lunch. 
We first provided an overview of the study and informal training, including 
listening to a variety of clip pairs and clip sets, making judgments, and discussing
the bases for judgments.  
(There was often variation, and from the discussions
it was clear that this was often due to noticing different 
aspects or weighting them differently. 
For example, for some clips some judges reported focusing 
more on similarity in terms of the ``feelings" conveyed,
while others focused more on similarity in terms of the ``sound".)
For each of the 31 sets, judges listened to the ten pairs, 
each of which including both the reference and the candidate, separated by a beep. 
They rated each pair using QuestionPro sliders.
Any judge could request repetitions of any pairs until they felt confident in their ratings.
Typically over half of the pairs were re-listened to in this way, often multiple times. The judges were all in one room, so all heard the same stimuli the same number of times. 

\begin{table}
    \begin{center}
        \begin{tabular}{l|l|l|l}
          & Random   & Our & Human \\
          & Baseline & Method & Judges \\
          \hline
          Ratings Correlation & 0.00 & 0.18 & 0.28\rule{0ex}{2.6ex} \\
          Recall@1 (= Precision@1) & ~~10\% &  ~~15\% & ~~20\% \\
          Recall@3 & ~~30\% & ~~43\% & ~~48\%\\
          Top 3 Intersection & 0.90 & 1.12 & 1.24
        \end{tabular}
    \end{center}
\caption{Ability to Identify the Most-Similar Pairs}
\label{tab:results}
\end{table}


\subsection{Preliminary Results}

Overall, a good model is one whose predicted most similar pairs
are actually rated highly similar by the human judges.  
Table \ref{tab:results} shows the results.  
First is the correlations between system and human ratings, 
across all 310 pairs and all 9 judges. Clearly this is a hard task even for human judges.
Further, we found that the model's ability to predict the {\it averages} of the human judgments, 
which we can take as an approximate gold standard, 
was much higher: 0.30.
We further examined whether the difference between the model's ratings
of two pairs could serve as a measure of its confidence,
and indeed there was a positive correlation between the difference
and the number of times the system correctly predicted human judgments
of which of the pairs was more similar.
We did not systematically explore the modeling space, but
we did obtain a better correlation by modifying Segura's model 
to normalize each feature to have 0 mean,
across the entire training data, before taking the cosine. 
However this modification hurt performance on the more important metrics.

The important metrics for the use cases are
seen in the next three lines of Table \ref{tab:results}.
First, we see that exact agreement on the \#1 most similar clip was low, 
both for the system and the humans.  
This was not surprising, because all of the top
3 were designed to be very close in similarity. 
Second, we see that the human-judged \#1 most similar clip was hard to  identify even
approximately, as it appeared in the system's top 3 choices only 43\% of the time, 
and humans did only slightly better. 
The last measure is the most relevant: 
the number of clips in the top-3 human rankings that appeared in the model's top-3 rankings, 
averaged over all 9 judges across all 31 sets. 
As seen in the table, the system's performance was above expectation,
and this was significant  ($p<.005$, t-test, 31 samples), but below human performance. 

To better understand the performance of the model, we did failure analysis
over two overlapping subsets of the data. 
The first was the pairs for which the system's predicted similarity diverged most
from the average human judgments. 
The second was the pairs that the model rated in the top 3, 
but whose average similarity ranking by the human judges was low, 
representing the failures that would be most hurtful for the use cases. 
We found three common causes. 
The first common cause was pairs for which one or both utterances were acoustically
unlike most of the training data.  
For example, these included an utterance with an ingressive exclamation,
and several cases which were very quiet and reduced, 
apparently spoken between friends who understood each other well.  
In most use cases, such unusual utterances will likely be rare, 
so this may not be problematic in practice. 
The second common cause was that the model was apparently often
sensitive to pragmatic differences that were 
less important to our judges, 
such as different turn-taking intentions (hold versus yield),
and the different evidentiality (statements based on first-person experience versus not). 
The third common cause was, conversely, that the model was sometimes
apparently sensitive
to dimensions of similarity that the human judges overlooked,
or which may have been overshadowed by salient category distinctions.
For example, this was the case for two clips, one in the form of a question 
and the other in the form of a statement,
which both were seeking a confirmation of an amount or quantity.

Overall we see that the model has some ability to pick out the  most-similar utterances, and no obvious show-stopper failure modes. 

\section{Classifying Speakers by Condition}
 
Hypothesizing that this ability is good enough to be useful,
we tested it on two instances of the classic problem of classifying
speakers as having a communicative disorder or not.

\subsection{Classification Method}

Our method is based on the assumption that the utterances of people
with a condition will be mostly similar to those of other people with
that condition, and the same for people without the condition.  
This assumption is
probably not unrealistic, unlike the assumption often
made, for many conditions, that there is a core underlying cause or marker.
For autism, for example, it is known that the prosodic behaviors
are diverse, as seen by conflicting results in the literature
\cite{mccann-peppe,mcalpine14,kiss-disso,wehrle19}, referencing both
inappropriately loud speech and too soft voice, and both monotone
pitch and overly wide pitch range.  Use of a similarity-based method
can be robust to such diversity.

Our algorithm has three steps.  First, it classifies each utterance of
an unknown speaker as likely having a condition or likely typically
developing based on the majority of the 7 nearest neighbors in the
reference data (kNN).  It does this 24 times, using all 24
layers of HuBert features in turn.  Second, it makes an overall
classification of each clip based on the majority vote of the
per-layer classifications.  Finally it classifies the speaker as
having a condition or not based on the label assigned to majority of
his or her utterances.
We tested this algorithm leave-one-speaker-out style on two datasets. 

\subsection{Evaluation for Autism}

We first used the NMSU  dataset of age-matched autistic (ASD) and neurotypical 
(NT) adolescents engaged in a find-the-difference task with a confederate
\cite{lehnert-fip20}.
For this data set there were 28 speakers, with typically 5-10 minutes
of data for each.  

As seen in Table \ref{tab:asd-nt}, the accuracy was
82\%, with 1 neurotypical and 4 autistic speakers misclassified.  
Lacking publicly-available data for evaluation, there is no way to
benchmark this method.  However these results do not
seem inferior to those reported by the most comparable previous work \cite{bone12,bone2015applying}. To
clarify, we do not claim this to be an advance in autism detection,
but rather a demonstration that a very simple model, whose parameters
are fixed and require no tuning on autism-specific data, still has
substantial discriminative power.

Further, upon examination, 
we found that the misclassified speakers were of three categories:
having inadequate data (few utterances, as they spoke less), being
among the youngest neurotypicals, or having autism diagnoses but
relatively low ADOS scores.
Thus the mispredictions do not seem to be attributable 
to failures of the similarity model or the kNN approach.

\begin{table}
    \begin{center}
        \begin{tabular}{l|rr}
          &\multicolumn{2}{c}{model prediction}  \\
          &  ASD & NT \\
          \hline
          speakers with autism diagnoses & 10 & 4 \\
          neurotypical speakers & 1 & 13
        \end{tabular}
    \end{center}
\caption{Autistic vs Neurotypical discrimination}
\label{tab:asd-nt}
\end{table}



\begin{table}
    \begin{center}
        \begin{tabular}{l|rr}
          &\multicolumn{2}{c}{model prediction}  \\
          &  SLI & TD \\
          \hline
          SLI speakers & 36 & 31 \\
          typically developing  speakers & 3 & 64
        \end{tabular}
    \end{center}
\caption{SLI vs TD discrimination}
\label{tab:sli-td}
\end{table}

\subsection{Evaluation for Specific Language Impairment}

We next used the Edmonton Narrative Norms Instrument
\cite{schneider-enni}, a corpus of children ages 4 to 10
retelling stories, served at Talkbank, including children with
Specific Language Impairment (SLI) \cite{sharma2021-sli},
and typically developing (TD) children.  SLI is not known to  
involve a pragmatics-related deficit, but we chose it because
of availability and our interest in data from younger children. 
The audio was segmented into utterances.  
As there was more data in the
typically-developing category, we downsampled to 67 speakers, matching
the age distribution of the SLI children.  There were an average of
about 35 utterances per child.
The results are fair, as seen in Table \ref{tab:sli-td}.
Further, examining dependencies on age
we found the discriminations to be at chance for the oldest group,
the 10-year olds, but fairly good for all other ages.
For comparison, we built a baseline using
the average length of utterance, in seconds, which
classified children whose average utterance length was less than 70\%
of the TD average as SLI.  Its performance was above chance, 
but the similarity-based method was far more accurate. 
Overall, this suggests that our model is capturing aspects
of similarity beyond the purely pragmatic.  
Depending on the purpose, this can be a good thing. 
\subsection{Performance without Condition-Specific Data} \label{sec:without}

A common low-resource scenario is one in which we have only data from
the typical/normal population, with no condition-specific data.  If we
assume that communicative disorders are departures from the norm, it
should be possible to identify an individual as impaired or not, even
without any disorder-specific data.

If this assumption is valid, then we can use a similarity
metric to detect atypicality: if many of a speaker's utterances
are not similar to anything produced by the reference/normal
population, then they may have some condition.  Of course, such a
vague characterization would not be useful for diagnosis, but could
be actionable for screening and referring children for
professional evaluation.

To test this idea, we again used the NMSU and ENNI data sets, 
with two measures of typicality:
an utterance-by-utterance model and a speaker-centroid model. 
In the former, we rated speaker typicality by how similar their utterances were, 
on average, to the closest 3 of those of all the typical speakers
(while leaving out the speaker himself, if in the typical set). 
In the second model, we rated speaker typicality by the distance between the 
centroid of all their utterances and the centroid of all typically-developing speakers.
For both models, we varied the thresholds post hoc, however 
performance was smooth, so this likely only slightly overstates the actual utility.
For the ENNI data there was no benefit with either model, but for the NMSU data both models did moderately well, with the centroid model doing slightly better.  In 
particular, a cosine threshold of 0.97 gave the performance shown in Table \ref{tab:asd2}.

\begin{table}
    \begin{center}
        \begin{tabular}{l|rr}
          &\multicolumn{2}{c}{model prediction}  \\
          &  ASD & NT \\
          \hline
          speakers with autism diagnoses & 13 & 8 \\
          neurotypical speakers & 1 & 6
        \end{tabular}
    \end{center}
\caption{ASD vs NT discrimination without ASD data}
\label{tab:asd2}
\end{table}

Unsurprisingly, the performance is not as good 
as when we exploit ASD-specific data. But we find some
value even without. We plan to investigate whether  this can be improved by conditioning on the context
and/or by excluding utterances that are atypical for the speaker.


\section{Contributions and Outlook}

Our first contribution is  an analysis of ways in which 
a general pragmatic-similarity model could be a useful addition
to the speech technology toolkit for communication disorders.  

Our second contribution 
is the finding that similarity can capture
utterance aspects that are relevant to diagnoses (Tables \ref{tab:asd-nt} and
\ref{tab:sli-td}).  Since nothing in our approach was
autism-specific or SLI-specific, we expect that this can serve as
a new approach useful for many other conditions, or mixes of conditions. 
We are seeking data to enable us to test this.

Our third contribution, regarding modeling using only small data (Need 2), 
is the finding that similarity can support discriminations without task-specific training
(Table \ref{tab:asd2}).  
This indicates the potential of shunting
off the hard modeling work to a general model, in line with the
increasingly popular strategy of pretraining on 
large, general data, and then adapting the model to specific needs.

Our fourth contribution, relating to richly-understandable categorization (Need 1), 
is the finding that a pragmatic similarity model
can support at least Use Case D: identifying most-similar utterances. 

Beyond these applied use cases, similarity models may also provide a new avenue for 
basic research on characterizing the space of communicative disorders.

We have obtained promising results despite using only a very
simple similarity model, that uses only the speech signal and only
considers single utterances.  Future work should seek to improve and
extend this: Beyond using the information in utterances, cases where
speakers say nothing, contrary to expectation, are also highly
informative.  Beyond the speech signal, a model might consider
multimodal information, such as gestures and activities.  Beyond
single utterances, a model might consider more context, such as the
interlocutor's recent behavior, or the recent interaction style
\cite{istyles}, as a proxy for the task and environment.  
Beyond acoustic-prosodic features, a model might also exploit the words.  
Beyond a black-box
similarity model, one could  build a fully explainable model
using perceptually-relevant prosodic features \cite{ward-marco-fuentes}.

Even with our existing model, much work remains to be done.
We need to build prototypes for the other use cases, obtain more data,
and test actual utility through user studies.

\medskip
\noindent
    {\bf Acknowledgments}

We thank Olac Fuentes for modeling advice, Arin Rahman for help with the analysis, and Pamela Hadley, Carol Anne Miller, Hedda Meadan-Kaplansky and Tracy Preza for discussion. 
This work was supported in part by the AI Research Institutes program of the National Science Foundation and the Institute of Education Sciences, U.S. Department of Education through Award \# 2229873 -- National AI Institute for Exceptional Education, and by NSF award 2348085.

\parskip 0pt

\bibliographystyle{ieeetr}
      \bibliography{bib}
     
\end{document}